\documentclass[11pt]{article}
\usepackage{cite}
\usepackage{amsmath,amssymb,amsfonts}
\usepackage{algorithmic}
\usepackage{hyperref}
\usepackage{graphicx}
\usepackage{textcomp}
\usepackage{xcolor}
\usepackage{comment}
\usepackage{INTERSPEECH_mod}
\usepackage{paralist}

\def\BibTeX{{\rm B\kern-.05em{\sc i\kern-.025em b}\kern-.08em
    T\kern-.1667em\lower.7ex\hbox{E}\kern-.125emX}}

\title{Modernizing Open-Set Speech Language Identification}

\name{Mustafa Eyceoz$^1$, Justin Lee$^1$, and Homayoon Beigi$^3$}
\address{
  $^{1,2}$Dept. of Computer Science, Columbia University, New York\\
  $^3$Recognition Technologies, Inc. and Columbia University, New York}
\email{$^1$me2680@columbia.edu, $^2$jjl2245@columbia.edu, $^3$beigi@recotechnologies.com}

\begin{document}

\maketitle

\begin{abstract}
While most modern speech Language Identification methods are
closed-set, we want to see if they can be modified and adapted for the
open-set problem. When switching to the open-set problem, the solution
gains the ability to reject an audio input when it fails to match any
of our known language options. We tackle the open-set task by adapting
two modern-day state-of-the-art approaches to closed-set language
identification: the first using a CRNN with attention and the second
using a TDNN. In addition to enhancing our input feature embeddings
using MFCCs, log spectral features, and pitch, we will be attempting
two approaches to out-of-set language detection: one using thresholds,
and the other essentially performing a verification task. We will
compare both the performance of the TDNN and the CRNN, as well as our
detection approaches.

\end{abstract}


\section{Introduction}
Speech Language Identification is the process of taking audio as an
input and determining what language is being spoken, if any. There are
two subsections to the language identification problem (which will
henceforth be referred to as LID): open-set and
closed-set~\cite{r:beigi-sr-book-2011}. In closed-set LID, set of
languages to identify is defined, and for every audio input, the "most
probable" language within the set is outputted. In open-set LID,
however, we also gain the option to "reject" that prediction and
detect when the audio input matches none of our known languages
well. It can also allow for identification and learning of new
languages for the system.

Today, there are a number of modern-day state-of-the-art approaches to
language identification, but almost all of them have opted to take the
closed-set approach. In an era of data abundance, the limitations of
the closed-set solution are typically circumvented by including
hundreds of languages and training on thousands of hours of data for
each of them. This workaround is obviously still not as ideal as the
true open-set solution, though, as it lacks the ability to detect and
reject or learn unknown languages, and in these cases it will
unavoidably output an incorrect prediction. Therefore, our goal is
attempt to adapt and modify these various state-of-the-art closed-set
solutions to attempt the open-set task, and see how well they perform,
as well as determine which implementation performs the best.

\section{Related Works and State of the Art}
To start, we will be looking at a few of the current best-performing
closed-set solutions. Convolutional Recurrent Neural Networks, or
CRNNs, have become increasingly popular in LID over recent
years. Solutions like that of Bartz et al in 2017
\cite{bartz2017language} initially used spectrograms as inputs, but
over a number of iterations we have seen the best performance come
from solutions like the recent 2021 paper from Mandal et al
\cite{mandal2021attention}, which proposes the use of a CRNN with
attention, as well as using Mel-frequency Cepstral Coefficient
(MFCC)~\cite{r:beigi-sr-book-2011} features of audio samples as input.

Another classic yet still high-performing method of both LID and
general speech recognition is to use the same MFCCs, but now use them
as input for a time-delay neural network, or TDNN. TDNN's are capable
of modeling long-term context information, which is why they are
often used in various speech recognition tasks.

There is also a third method of speech LID that, either in this paper
or in future works, may be worth exploring. This method actually
separates the tasks of speech recognition and language
identification. First, the speech is converted to text. While TDNNs
have long been used for speech-to-text, the current top performers for
this task are all wave2vec 2.0 \cite{baevski2020wav2vec}
implementations. Once the text has been obtained, textual LID is
currently done best using bi-directional LSTMs (long short-term
memory, a type of recurrent neural network or RNN architecture), like
the implementation by Toftrup et al \cite{toftrup2021reproduction}
that builds off an outline by Apple.

For this paper, we primarily focus on the direct speech methods for
LID, modifying both the input selection and various other points of
the architectures (to be touched on further in section IV. Proposed
Methodology). It is also possible that we explore the speech-text-LID
method as well, though that may be left for a a continuation of this
work.

When looking at previous open-set work, we draw inspiration from both
the exploration of thresholding functions and curves for out-of-set
language detection and rejection, like the work of Rebai et al
\cite{8308370}, as well as creating deeper embeddings with linear
discriminant analysis (LDA)~\cite{r:beigi-sr-book-2011} transformation
sets to perform verification tasks, like with Voxceleb2 speaker
verification \cite{2018}.

\section{Datasets}
The dataset used in this research consists of audio and text from 9
different language sources. For our in-set languages, we will be using
French, Turkish, Spanish, Korean, Mandarin, English, and Russian. And
for our out-of-set languages, we will be evaluating using Javanese and
Bengali.

MediaSpeech is a dataset containing French, Turkish and Spanish media
speech. Originally built with the purpose of testing Automated Speech
Recognition (ASR) systems performance, MediaSpeech contains 10 hours
of speech for each language provided. \cite{mediaspeech2021}
Pansori-TEDxKR is a dataset that is generated from Korean language
TEDx talks from 2010 to 2014. This corpus has about three hours of
speech from 41 speakers. \cite{pansori} Primewords Chinese Corpus Set
1 is a Chinese Mandarin corpus released by Shanghai Primewords
Co. Ltd. and contains 100 hours of speech. This corpus was built from
smart phone recordings of 296 native Chinese speakers and has
transcription accuracy of larger than 98\% at a confidence level of
95\%. \cite{primewords_201801} Free ST American English Corpus is a
free American English corpus by Surfingtech. It contains the
utterances of 10 speakers with each speaker having approximately 350
utterances. \cite{freest_english} Russian LibriSpeech is a Russian
dataset based on LibriVox audiobooks. It contains approximately 98
hours of audio data. \cite{russian_librispeech} Note that each of the
datasets mentioned above will be normalized so that each in-set
language is represented by an equal number of hours of audio in order
to prevent any skewing in the in-set languages.

For evaluation, the first additional out-of-set languages is
Javanese. The Large Javanese ASR training data set contains
approximately 185,000 utterances in Javanese and was collected by
Google in collaboration with Reykjavik University and Universitas
Gadjah Mada in Indonesia. \cite{kjartansson-etal-sltu2018} The second
out-of-set language is Bengali. The Large Bengali ASR training data
set contains approximately 196,000 utterances in Bengali and contains
transcribed audio data for Bengali. \cite{kjartansson-etal-sltu2018}

\section{Proposed Methodology}
The two methods we will be adapting and comparing for open-set
performance are the CRNN with attention solution and the TDNN
solution.

First, to obtain our feature embeddings to use as input for the TDNN
and CRNN, we must process the data through a number of steps. We start
by performing a discrete Fourier transform on data frames to generate
the spectral estimates. From this we can obtain the log spectral
features. With an additional discrete cosine transform we can obtain
the MFCCs. We then concatenate the log spectral
features~\cite{r:beigi-sr-book-2011} with the MFCCs, as well as some
additional pitch information. To ensure our embeddings have all the
information needed for the task, we may also concatenate 100
dimensional i-vectors~\cite{r-m:dehak-2009-1}. From here, we will then
pass these embeddings through an LDA~\cite{r:beigi-sr-book-2011} to
perform both dimensionality and correlation reduction of the features.

Once we have obtained our final feature embeddings, we will be using
them to train and compare our two models: the first being our CRNN
with attention, and the second being our TDNN. For both, our initial
implementation will have a softmax output layer for language
identification, as well as threshold curves and functions used on
output for out-of-set language detection and rejection. Our second
approach will be more complex. After training our softmax output, we
will attempt to continue to a deeper embedding by training an LDA
transformation set, both allowing us to treat the open-set problem as
a verification task, as well as potentially giving us the ability to
add new languages without having to retrain the initial model, instead
simply using these deeper embeddings.

We will be comparing performance of the TDNN model vs the CRNN model
on both of the proposed approaches, as well as seeing which of the two
approaches generally performs better on the open-set task. We
currently expect the more modern CRNN model to at least slightly
outperform the TDNN model, though we have no expectation for which of
the out-of-set detection approaches will perform better (although
regardless the verification approach will provide additional
functionality over the threshold approach).

It is worth noting that there is also a chance we attempt these two
approaches on the word2vec 2.0 + bi-LSTM method of LID, but that may
very well be saved for future work.

\begin{figure}[htb]
\begin{minipage}[b]{1.0\linewidth}
\centerline{\includegraphics[scale=0.61]{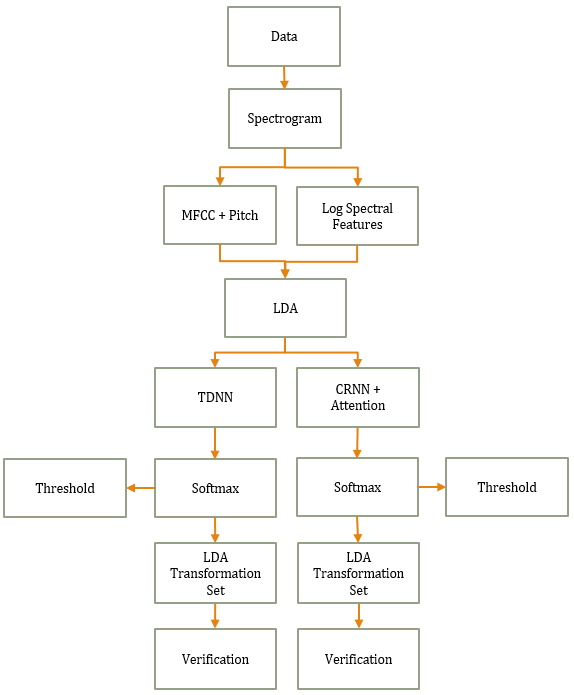}}
\caption{Open-Set LID Exploration Architecture}
\label{fig:openset_lid_architecture} 
\end{minipage}
\end{figure}

See Figure~\ref{fig:openset_lid_architecture} for the proposed
architecture for our open-set language identification approach.

\section{The Process: Data Preparation and Feature Extraction}
\subsection{Data Preparation}
After downloading the necessary datasets from Open Speech and Language
Resources (OpenSLR) \cite{openslr}, we proceeded with data preparation
so that all of the data for each language was in the correct format
and structure for the Kaldi scripts used in feature extraction. We
started by formatting each dataset in the same way: all datasets came
with different file formats and structures; and some came with audio
data in the form of FLAC files while others came as WAV files.

The first step in our data formatting process was to convert all audio
files to WAV format. WAV, also known as Waveform Audio File Format, is
the main format of audio that is used by the Kaldi scripts, which are
subsequently used for feature extraction. Each audio file in our
original datasets were either FLAC or WAV, so all FLAC files were
converted to WAV using ffmpeg software.

We measured the total duration of data we had for each language and
then limited each language's data to 10 hours, as that was the minimum
total duration of audio files found across all languages. For some
languages, we had up to 40 hours of data, but the reason for using an
equal duration of audio files for each language was to reduce any
skewing towards certain languages when training. This was all done
using librosa, a python library for audio analysis.

Then, the data in each language was split with an 80:20 train and test
split. It's worth noting that the split was made based on the total
duration of the audio files, not the number of audio files
present. From here, several acoustic data files were generated. First,
'wav.scp' files were created for each data split of each language
which contain data that maps a WAV file's unique identifier to its
file path. Then, a 'text' file was created for each data split of each
language which contains a map of every audio file to its text
transcription. We then created the 'corpus.txt' files for each
language which contained every single utterance transcription from the
audio files of said language. Finally other extraneous files were
created such as 'utt2spk' which, for our use-case, mapped a WAV file's
unique identifier to itself since our dataset and problem statement
does not involve individual speakers.

Then, we created several files related to language data as
follows. For each language and their transcriptions of the audio
files, the 1000 most frequent words were computed and saved. These
1000 most frequent words represent the dictionary of a language and
the most significant identifiers for that language. We used these to
create a 'lexicon.txt' for each language, which contains all of the
1000 most frequent words with their phone transcriptions. Since we
could not find the necessary tools to convert words into phones for
all 9 languages, we resorted to the solution of using each individual
letter as a phone, also known as graphemic transcription. One point of
concern came when working with Mandarin, in which each character is a
pictorial representation of a word and therefore has no concept of
letters. Thus, the solution was to first convert Mandarin into Pinyin,
which is the romanticized text version of Mandarin. From here, it was
easy to split a Pinyin word into individual letters. With all of the
aforementioned phones, we combined them with the silence phones to
create 'lexicon.txt'. Then, it was straightforward to create
individual 'nonsilence\_phones.txt' and 'silence\_phones.txt' files
which contain the non-silent phones and silent phones
respectively. The silent phones are 'sil' and 'spn'. Finally, the
'optional\_silence.txt' file was created with just the phone 'sil'.

\subsection{Feature Extraction}
From here, data preparation was completed and we moved to feature
extraction. We used the Kaldi script 'make\_mfcc\_pitch.sh' to extract
the Mel-frequency cepstral coefficients
(MFCC)~\cite{r:beigi-sr-book-2011} features and pitch data for each
audio file. Then, the Mel-spectral
features~\cite{r:beigi-sr-book-2011} were extracted from the audio
files using the python library librosa. From here, we used another
python library called Kaldiio to read the ark files that contain the
MFCC and pitch data. For the final feature embeddings, all
aforementioned features were concatenated and passed through Linear
Discriminant Analysis (LDA)~\cite{r:beigi-sr-book-2011} in order to
perform both dimensionality and correlation reduction of the features.

\section{The Process: Rival Models}
\subsection{Convolutional Recurrent Neural Network with Attention}
Our first model is a Convolutional Recurrent Neural Network (CRNN)
with attention. CRNNs are essentially a Convolutional Neural Network
(CNN) followed by a Recurrent Neural Network (RNN). Our CRNN approach
uses a 2-dimensional CNN and then an RNN built with Bidirectional Long
short-term memory (BiLSTM) layers.

Specifically, our model contains two 2-dimensional convolution layers
both with kernel size of 2 and whose outputs are flattened and
concatenated to the original feature embeddings. This is then all
passed through two BiLSTM layers both with 256 recurrent layers and
256 hidden features. Then, we add attention which allows the model to
focus on the relationship between different discriminative
features. The attention mechanism is encapsulated as a single layer
that comes after the two BiLSTM layers and has output size of 7 for
the 7 in-set languages. \cite{crnnagithub} We then add a softmax
output layer that would typically be used for language identification
since it allows us to make a language prediction with highest
probability. Finally, to adapt this model to the open-set language
identification problem, a threshold is used so that if all of the
probabilities outputted by the softmax layer are under this threshold,
the input is deemed out of the set and is rejected.

Other details related to this model include the loss function and
optimizer, for which we used cross-entropy loss and stochastic
gradient descent respectively. This model with the aforementioned loss
function and optimizer was trained for 12 epochs.

\begin{figure}[htb]
\begin{minipage}[b]{1.0\linewidth}
\centerline{\includegraphics[scale=0.45]{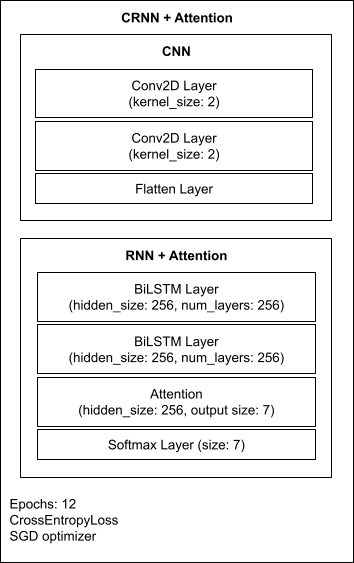}}
\caption{CRNN + Attention Architecture}
\label{fig:crnn_architecture} 
\end{minipage}
\end{figure}

See Figure~\ref{fig:crnn_architecture} for a diagram of the CRNN +
attention architecture.

\subsection{Time Delay Neural Network}
Our second model is a Time delay neural network (TDNN). TDNNs are
feed-forward networks that can model long-term context
information. TDNNs are especially good at modeling context and
classifying patterns without needing any explicit segmentation of
input data. The key hyper-parameters of each layer in a TDNN are
context size, dilation, and stride which describe the number of
contiguous frames to observe, number of non-contiguous frames to
observe, and how many frames to skip in an iteration.

Specifically, our model contains six layers of sizes 512, 512, 512,
512, 1500, 7 with context sizes of 5, 3, 3, 1, 1, 1 respectively. Each
layer has dilation of 1, 2, 3, 1, 1, 1 respectively. All layers have
stride of 1 and uses the ReLU activation function. \cite{tdnngithub}
We then add a final softmax output layer that would typically be used
for language identification since it allows us to make a language
prediction with highest probability. Finally, to adapt this model to
the open-set language identification problem, a threshold is used so
that if all of the probabilities outputted by the softmax layer are
under this threshold, the input is deemed out of the set and is
rejected.

Other details related to this model include the loss function and
optimizer, for which we used cross-entropy loss and stochastic
gradient descent respectively. This model with the aforementioned loss
function and optimizer was trained for 12 epochs.

\begin{figure}[htb]
\begin{minipage}[b]{1.0\linewidth}
\centerline{\includegraphics[scale=0.35]{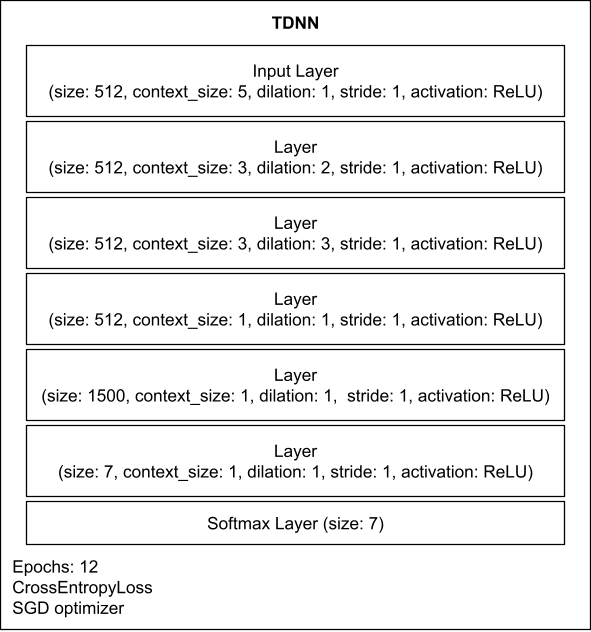}}
\caption{TDNN Architecture}
\label{fig:tdnn_architercture} 
\end{minipage}
\end{figure}

See Figure~\ref{fig:tdnn_architercture} for a diagram of the TDNN
architecture.

\section{Results}

\subsection{Convolutional Recurrent Neural Network with Attention}
The training data used when training our CRNN with attention included
80\% of the total duration of audio files from each of the 7 in-set
languages: English, Spanish, French, Korean, Mandarin, Russian, and
Turkish. After training the model for 12 epochs on this training data
and before experimenting with any thresholds, we first tested it on
the other 20\% of the total duration of audio files from just the 7
in-set languages again in order to gauge how well our model performs
at the closed-set language identification task. The CRNN with
attention was able to achieved an in-set accuracy of 85\%; that is,
the model was able to correctly identify the language of a given
in-set language audio input 85\% of the time.

We then incorporated a threshold as the final layer of the model and
tested the CRNN with attention on the other 20\% of the total duration
of audio files from the 7 in-set languages as well as the 2 out-of-set
languages: Javanese and Bengali. We measured three accuracies: overall
accuracy, in-set accuracy, and out-of-set accuracy. The overall
accuracy is how often our model was able to take an input and
correctly identify its language or reject it if it was not one of the
7 in-set languages. The in-set accuracy describes how often our model
was able to correctly identify the language of an in-set audio
input. The out-of-set accuracy describes how often it was able to
correctly reject an out-of-set audio input.

\begin{figure}[htb]
\begin{minipage}[b]{1.0\linewidth}
\centerline{\includegraphics[scale=0.55]{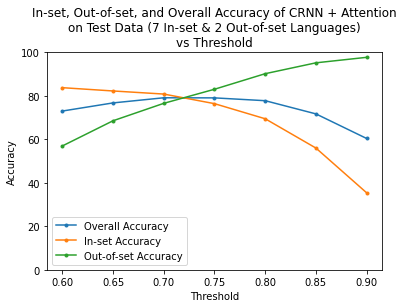}}
\caption{CRNN + Attention Accuracy (\%) Breakdown}
\label{fig:crnn_accuracy_breakdown} 
\end{minipage}
\end{figure}

We noticed that the overall accuracy of our model reached a maximum of
0.791 across the various thresholds we experimented with. With too
small of a threshold, the overall accuracy was about 70\% but as the
threshold was increased slightly, the overall accuracy began to
increase. At around a threshold of 0.7 and 0.75, the overall accuracy
started to reach its maximum. However, when the threshold was set to
be too high, the overall accuracy dropped drastically as too high of a
threshold resulted in perfect out-of-set accuracy but terrible in-set
accuracy. Using a threshold of 0.7, the CRNN with attention achieves
the maximal overall accuracy of 79.1\% along with in-set and out-of-set
accuracies of 80.8\% and 76.6\% respectively. See
Figure~\ref{fig:crnn_accuracy_breakdown} for a plot of the overall
accuracy for various thresholds and Table~\ref{tab:crnn_accuracy} for
the exact data.

Compared to the state of the art for closed-set (the 2021 paper from
Mandal et al \cite{mandal2021attention}) which uses their own CRNN
with attention model and has an accuracy of 98\% (and 91\% in noisy
environments), our CRNN with attention model has an in-set language
identification accuracy of 85\%. CNNs, which are a core part of our
CRNN model, are mainly used for image classification. Since the CNN is
able to model local context connectivity, it is generally useful for
tasks with images where a sliding filter would be practical. In our
feature embedding, which uses MFCC and Mel-Spectral
features~\cite{r:beigi-sr-book-2011}, there is little connectivity
between dimensions, which is an argument for why our 2-dimensional CNN
implementation may not have been optimal. However, since there is
still local context connectivity across time steps in our data, a
1-dimensional CNN at the start of our CRNN with attention
implementation may have been a better choice.

\begin{table}[htb]
\begin{minipage}[b]{1.0\linewidth}
  \centering
  \begin{tabular}{c|c|c|c}
    \hline
    Threshold & \multicolumn{3}{c}{Accuracy (\%)} \\
    \hline
         & Overall & In-set & Out-of-set\\
        0.6 & 73.0 & 83.8 & 56.9 \\
        0.65 & 76.8 & 82.2 & 68.5 \\
        0.7 & 79.1 & 80.8 & 76.6 \\
        0.75 & 79.1 & 76.4 & 83.0 \\
        0.8 & 77.8 & 69.5 & 90.2 \\
        0.85 & 71.7 & 56.0 & 95.2 \\
        0.9 & 60.4 & 35.4 & 97.7 \\
        \hline
  \end{tabular}
  \caption{Accuracy Breakdown of CRNN + Attention vs Threshold}
  \label{tab:crnn_accuracy}
\end{minipage}
\end{table}

\subsection{Time Delay Neural Network}
The training data used when training our TDNN is identical to the
training data of the first model. After training the model for 12
epochs on this training data and before experimenting with any
thresholds, we first tested it on the other 20\% of the total duration
of audio files from just the 7 in-set languages again in order to
gauge how well our model performs at the closed-set language
identification task. The TDNN was able to achieved an in-set accuracy
of 95\%; that is, the TDNN was able to correctly identify the language
of a given in-set language audio input 95\% of the time.

While this assuredly beats out our CRNN + attention model, when
comparing to the state of the art CRNN + attention with accuracy of
98\%, our TDNN model with an in-set language identification accuracy
of 95\% still falls just a bit under. It is worth noting, however,
that this is with training/testing on only 70 hours of speech data,
which gives us hope that the TDNN model could actually rival state of
the art CRNNs.

\begin{figure}[htb]
\begin{minipage}[b]{1.0\linewidth}
\centerline{\includegraphics[scale=0.55]{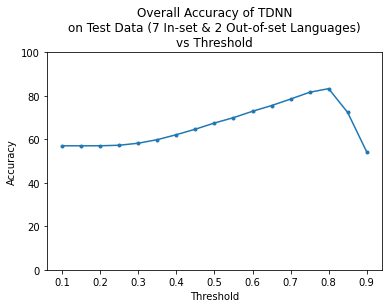}}
\caption{TDNN Overall Accuracy (\%)}
\label{fig:tdnn_overall_accuracy} 
\end{minipage}
\end{figure}

We then incorporated a threshold as the final layer of the model and
tested the TDNN on the other 20\% of the total duration of audio files
from the 7 in-set languages as well as the 2 out-of-set languages:
Javanese and Bengali. First, we measured our model's overall
accuracy. We noticed that the overall accuracy of our model reached a
maximum of 83\% across the various thresholds we experimented
with. With too small of a threshold, the overall accuracy was about
60\% but as the threshold was increased slightly, the overall accuracy
began to increase. At around a threshold of 0.7 and 0.8, the overall
accuracy started to reach its maximum. However, when the threshold was
set to be too high, the overall accuracy dropped as too high of a
threshold resulted in perfect out-of-set accuracy but terrible in-set
accuracy. See Figure~\ref{fig:tdnn_overall_accuracy} for a plot of the
overall accuracy for various thresholds and
Table~\ref{tab:overall_tdnn_accuracy} for the exact data.

\begin{figure}[htb]
\begin{minipage}[b]{1.0\linewidth}
\centerline{\includegraphics[scale=0.55]{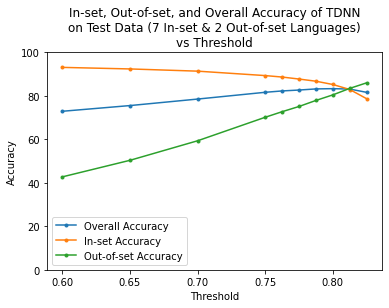}}
\caption{TDNN Accuracy (\%) Breakdown}
\label{fig:tdnn_accuracy_breakdown} 
\end{minipage}
\end{figure}

Taking a few of the best threshold results from the previous step, we
took a closer look into the in-set and out-of-set accuracies
individually. See Figure~\ref{fig:tdnn_accuracy_breakdown} for a
breakdown of the accuracy of the TDNN and
Table~\ref{tab:tdnn_accuracy_breakdown} for the exact data. At about a
threshold of 0.8, we achieve the maximal overall accuracy of 83.3\% and
in-set and out-of-set accuracies of 85.2\% and 80.4\% respectively.

\begin{table}[htb]
\begin{minipage}[b]{1.0\linewidth}
  \centering
  \begin{tabular}{c|c}
    \hline
    Threshold     & Overall Accuracy (\%)\\
    \hline
    0.1 & 57.0\\
    0.15 & 57.0\\
    0.2 & 57.0\\
    0.25 & 57.2\\
    0.3 & 58.2\\
    0.35 & 59.8\\
    0.4 & 62.1\\
    0.45 & 64.6\\
    0.5 & 67.5\\
    0.55 & 69.9\\
    0.6 & 72.9\\
    0.65 & 75.5\\
    0.7 & 78.5\\
    0.75 & 81.6\\
    0.8 & 83.3\\
    0.85 & 72.3\\
    0.9 & 54.2\\
    \hline
  \end{tabular}
  \caption{Overall Accuracy of TDNN vs Threshold}
  \label{tab:overall_tdnn_accuracy}
\end{minipage}
\end{table}

\begin{table}[htb]
\begin{minipage}[b]{1.0\linewidth}
  \centering
  \begin{tabular}{c|c|c|c}
    \hline
    Threshold & \multicolumn{3}{c}{Accuracy (\%)} \\
    \hline
         & Overall & In-set & Out-of-set \\
    0.6 & 72.9 & 93.1 & 42.7 \\
    0.65 & 75.5 & 92.4 & 50.4 \\
    0.7 & 78.5 & 91.4 & 59.3 \\
    0.75 & 81.6 & 89.3 & 70.1 \\
    0.7625 & 82.3 & 88.6 & 72.7 \\
    0.775 & 82.7 & 87.7 & 75.1 \\
    0.7875 & 83.2 & 86.7 & 77.9 \\
    0.8 & 83.3 & 85.2 & 80.4 \\
    0.8125 & 83.1 & 82.9 & 83.4 \\
    0.825 & 81.6 & 78.7 & 86.0\\
    \hline
  \end{tabular}
  \caption{Accuracy Breakdown of TDNN vs Threshold}
  \label{tab:tdnn_accuracy_breakdown}
\end{minipage}
\end{table}

\section{Conclusion and Future Work}
It is hard to come to any significant conclusions with respect to our
CRNN + attention model, as we were unable to reproduce
state-of-the-art performance on the initial closed-set task, meaning
our open-set results are likely also markedly worse than what the best
models could theoretically produce. This, however, is not to say that
our labor was without fruit. What we did learn was that even with a
modestly sized data set, a TDNN is still able to produce respectable
closed-set language identification results, and has the potential
(perhaps simply with more data) to rival the modern CRNNs that seem to
be dominating the field at the moment. Essentially, TDNNs should not
be counted out as an option when architecting modern solutions to
language identification tasks, and they could yet be the key to
breaking our existing barriers in the field. Furthermore, when
incorporating thresholds, the TDNN was still able to hold its own on
in-set data, even when pushing towards high out-of-set detection
accuracy.

There is still much to be done in terms of these experiments,
though. First and foremost is that a more diverse array of
thresholding and detection methods need to be tested. Surely modeling
a curve and picking the best static threshold is not the optimal
thresholding solution, and dynamic solutions (such as per-language
thresholds), could yield significantly improved results, perhaps
allowing us to retain an in-set accuracy much closer to our initial
95\%. Beyond this, one may notice that we did not get a chance to
experiment further with LDA transformation sets, and treating
out-of-set detection as a verification task. This will be an
important, if not necessary, step should one choose to continue on
this path of research. This is not only due to potential accuracy
gains, but also the functional advantages that these deeper embeddings
may provide, as mentioned in our initial reasoning.

Additionally, more can be done with respect to feature abstraction and
embedding. Further experimentation with respect to optimal
incorporation of Mel-Spectral features, as well as the
initially-planned i-vectors, could do a great deal in increasing even
our models' closed-set accuracies. Obviously there is much more work
to be done with our CRNN + attention model as well in getting it up to
modern standards, though our first course of action would likely be to
attempt to break the state of the art with our highly-performant TDNN,
and from there begin experimenting with other open-set detection
solutions to truly build something practically meaningful. Overall,
though, we have gained a great deal of knowledge and experience from
this experiment already, and look forward to attempting to take it
further.

\bibliographystyle{IEEEtran}
\bibliography{ms.bib} 

\end{document}